# A Distributed Approach to Meteorological Predictions: Addressing Data Imbalance in Precipitation Prediction Models through Federated Learning and GANs

Elaheh Jafarigol[1*] and Theodore B. Trafalis[2]

[1*]Data Science and Analytics Institute, University of Oklahoma, 202 W. Boyd St., Room 409, Norman, 73019, Ok, USA.
[2]Industrial and Systems Engineering, University of Oklahoma, 202 W. Boyd St., Room 104, Norman, 73019, OK, USA.

*Corresponding author(s). E-mail(s): elaheh.jafarigol@ou.edu;

**Abstract**

The classification of weather data involves categorizing meteorological phenomena into classes, thereby facilitating nuanced analyses and precise predictions for various sectors such as agriculture, aviation, and disaster management. This involves utilizing machine learning models to analyze large, multidimensional weather datasets for patterns and trends. These datasets may include variables such as temperature, humidity, wind speed, and pressure, contributing to meteorological conditions. Furthermore, it's imperative that classification algorithms proficiently navigate challenges such as data imbalances, where certain weather events (e.g., storms or extreme temperatures) might be underrepresented. This empirical study explores data augmentation methods to address imbalanced classes in tabular weather data in centralized and federated settings. Employing data augmentation techniques such as the Synthetic Minority Over-sampling Technique or Generative Adversarial Networks can improve the model's accuracy in classifying rare but critical weather events. Moreover, with advancements in federated learning, machine learning models can be trained across decentralized databases, ensuring privacy and data integrity while mitigating the need for centralized data storage and processing. Thus, the classification of weather data stands as a critical bridge, linking raw meteorological data to actionable insights, enhancing our capacity to anticipate and prepare for diverse weather conditions.

**Keywords:** Imbalanced learning, Federated learning, Deep learning, Generative Adversarial Networks, Weather prediction

## 1 Introduction

Employing federated learning for the classification of weather data introduces a new paradigm, particularly in scenarios where data privacy, decentralized data sources, and efficient utilization of localized data are critical. Local meteorology stations frequently collect weather data from measurements and radar observations in tabular and image formats.

Federated learning enables models to be trained directly on local devices or stations where the data resides, eliminating the need to transmit sensitive or voluminous data to a central location. Federated learning also allows private data to be monitored and protected by local data centers. Each local model learns from its respective data. Then, only the model updates (not the data) are shared with a global model, ensuring data privacy and reducing communication costs. The applications of federated learning have been extended to weather forecasting and air quality control using historical data and edge devices [1, 2]. This collaborative yet decentralized learning method is crucial for weather prediction due to the inherently localized nature of weather events and the potential sensitivity of data. Machine learning has long been used in weather applications to predict weather conditions such as rain or strong winds [3] to improve lead time for severe weather warnings, such as tornadoes [4]. Federated learning provides the model with a diverse and comprehensive dataset acquired from various local environments and conditions.

Given Australia's climatic and geographical diversity, the weather stations record a broad spectrum of meteorological patterns, providing models with insight into various scales and types of weather phenomena. Precise and localized weather forecasting is crucial for various applications, from agriculture to urban planning. To this end, we conduct an experimental study focusing on predicting rainy versus non-rainy days through deep learning models. The dichotomy of rainy and non-rainy days establishes a clear classification problem where the model is trained to discern the atmospheric variables that result in precipitation. The real-time system collects the data for this study at the Bureau of Meteorology in Australia. The dataset consists of 140672 observations and 13 features stored in 9 stations. The data contains approximately ten years of daily observations from 2007 to 2017, recorded twice daily from the eight mainland regions and Australian offshore islands. Figure 1 demonstrates the 20-year average rainfall measured annually in the eight major regions across Australia.

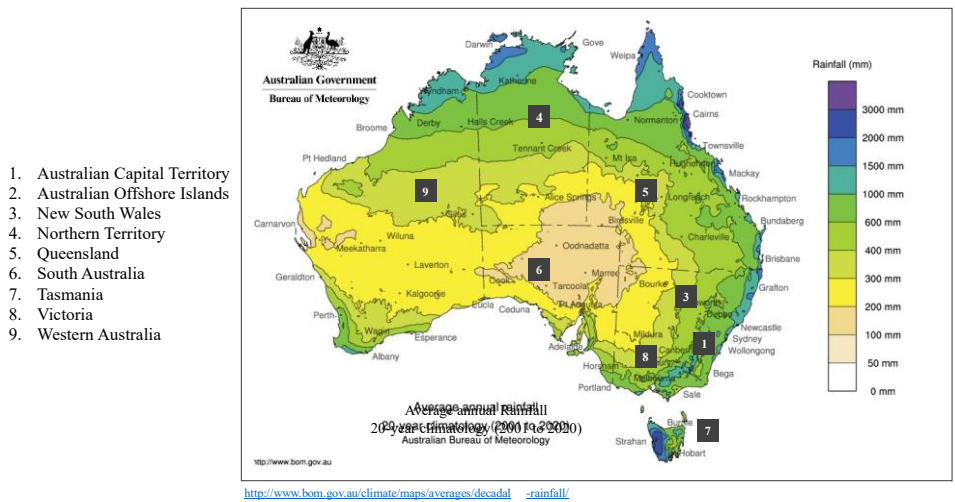

1. Australian Capital Territory
2. Australian Offshore Islands
3. New South Wales
4. Northern Territory
5. Queensland
6. South Australia
7. Tasmania
8. Victoria
9. Western Australia

http://www.bom.gov.au/climate/maps/averages/decadal-rainfall/

**Fig. 1**: Multi-decadal rainfall averages map presents the rainfall patterns across the regions in a 20-year period.

The challenges in the classification problem arise especially in regions where rainy days are sparse or seasonally confined. Central and southern regions have experienced less rainfall than northern and eastern coastal areas, resulting in varying ratios of rainy versus non-rainy days between the regions. With geographical and temporal variations in weather patterns, especially across a diverse continent like Australia, the issue of imbalanced learning is more likely to present itself in the significant difference in the number of observations in each class. Table 1 presents the distribution of observations in both datasets distinguished by the class label.

The two major challenges drawn from the information presented in Table 1 are:

| Station | Region | Rain | No Rain | Imbalance Ratio |
|---|---|---|---|---|
| 1 | Australian Capital Territory | 2016 | 7307 | 0.22 |
| 2 | Australia Offshore Islands | 919 | 2045 | 0.31 |
| 3 | New South Wales | 9305 | 32027 | 0.23 |
| 4 | Northern Territory | 1361 | 6421 | 0.17 |
| 5 | Queensland | 3513 | 11600 | 0.23 |
| 6 | South Australia | 2402 | 9710 | 0.20 |
| 7 | Tasmania | 1460 | 4756 | 0.23 |
| 8 | Victoria | 7217 | 23814 | 0.23 |
| 9 | Western Australia | 3568 | 11231 | 0.24 |

**Table 1**: Distribution of rain and no-rain observations across nine weather stations in Australia, indicating a data imbalance in the regional data.

1. The significant difference between the number of instances in the classes indicates that the data is imbalanced. Imbalanced data negatively impacts the classifier's performance, resulting in biased predictions.
2. Insufficient data in some local centers negatively affects the accuracy of predictions, resulting in models incapable of generalizing to new and unseen data.

Imbalanced data is a well-known issue in many fields, including weather prediction, and it is an ongoing topic in machine learning research[5, 6]. The relative proportion of classes and the absolute number of available instances in the minority class are important factors. The problem with imbalanced data is magnified when the minority class consists of rare events because there is a lack of general information on the event, leading to biased models. Tornadoes and thunderstorms happen at various frequencies in locations with different climate conditions. The rarity of such events creates imbalances in the data, which requires specialized methods to address this issue. Trafalis et al. [7] proposed a weighted classifier with a random subspace ensemble method to classify tornadic and non-tornadic observations. Predicting the intensity of the damages caused by a tornado is also a challenging problem [8]. When the data is imbalanced, machine learning classifiers fail to learn the underlying patterns within the minority class. Without a significant loss in overall accuracy, the minority class is misclassified. Based on the type of data, the size, and the distribution of the data between classes, the issue can affect the performance in different ways. Cost-sensitive methods are a practical approach to addressing the issue of imbalanced weather data. Jafarigol and Trafalis [9] proposed a novel linear programming Support Vector Machine that outperforms traditional machine learning algorithms in classifying weather data. A lack of adequate information about the minority class causes the problem definition issues [10]. This can cause evaluation metrics such as accuracy and error rate to fail in representing the minority class. Evaluation is an essential part of the learning process, which is used to assess the generalization ability of the learning method on test data. Appropriate evaluation metrics are necessary for evaluating the quality of learning [11–13]. The authors of the paper published by Ferri et al. [14] have used experimental and theoretical analysis to compare and rank the evaluation metrics that work best in evaluating the learned model on imbalanced data and analyze the identifiable clusters and relationships between the metrics. These experiments provide recommendations on the metrics that would be more appropriate for any specific application. Evaluation metrics are categorized into three types in the literature: threshold, probability, and ranking metrics [15]. The threshold evaluation metrics are computed based on the confusion matrix. In binary classification, given that samples in the majority class are labeled negative and the samples in the minority class are labeled positive, the confusion matrix is defined based on four values of True Positive(TP), True Negative(TN), False Positive(FP), and False Negative(FN) calculated based on the actual and predicted values. Note that the definition of a confusion matrix can be extended to multi-class classification. Figure 2 provides an overview of the challenges in imbalanced learning and the approaches that have led to efficacious solutions in this domain. Accuracy is limited to measuring the overall performance, and it cannot provide enough information to ensure a reliable learning method when the data is imbalanced [16]. *Sensitivity* and *Specificity* are two classification performance metrics for imbalanced learning. Sensitivity is $\frac{TP}{TP+FN}$ and summarizes how well the positive class was predicted. Specificity is defined as $\frac{TN}{TN+FP}$, and it evaluates how well the negative class was predicted. Geometric mean (G-mean) is an important evaluation metric used explicitly for imbalanced learning scenarios. G-mean considers the harmonic mean of sensitivity and specificity. A high G-mean indicates that the model performs well in both classes, so we aim to maximize the metrics. The goal of imbalanced learning is to find an optimal classifier that is capable of providing a balanced degree of predictive accuracy for the minority class as well as the majority class [17–22]. As shown in Figure 2, the Receiver Operating Characteristic curve visually represents the classification performance. The Area Under the Curve (AUC) is scale-invariant, so it is a reliable tool for the ranking and comparison of classifiers[23, 24].

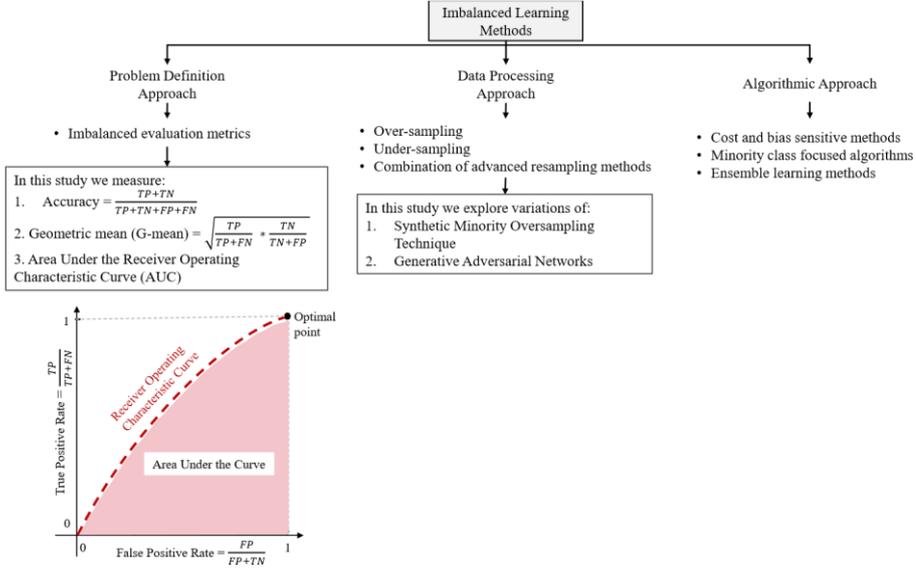

**Fig. 2**: Utilizing the appropriate imbalance learning evaluation metrics is a standard practice. In this study, accuracy, AUC, and G-mean are the key tools in the assessment and comparison of the oversampling techniques in both classes.

Motivated by these challenges, we empirically analyze data augmentation methods to balance the data prior to training in federated learning frameworks. Utilizing data augmentation techniques allows the generation of synthetic data points that mimic the characteristics of actual rainy days, thereby alleviating data scarcity and imbalance issues. Federated learning enables us to leverage the data from multiple centers without accumulating the data in a single facility. Moreover, in the federated learning context, each station can augment its data locally, ensuring that the synthetic data reflects the local meteorological characteristics and improving the learning of the localized model. We compare the Synthetic Minority Over-sampling Technique (SMOTE), a widely used approach, with Generative Adversarial Learning (GANs) variants. When the local models communicate and contribute to the training of a global model, the model's capability to generalize and accurately predict rainy events improves. This is especially true when the models are trained on the diversified and balanced representation of rainy days across different Australian climates and territories. Therefore, the meteorological predictions become more reliable and representative of the vast and varied Australian landscapes. In this paper, We train a deep learning model on a combination of real and synthetic data generated by various methods. With its ability to model complex, nonlinear relationships and learn hierarchical features from data, deep learning emerges as a quintessential tool in analyzing meteorological data and uncovering the underlying patterns that cause rainfall. Variables such as humidity, pressure, temperature gradients, and wind patterns are fed into the neural network. The model continuously refines its predictive capability through layered architectures and back-propagation, resulting in a more adaptive and accurate system capable of effectively distinguishing between rainy and non-rainy days. This predictive paradigm enhances meteorological forecasting and provides different sectors with insights to develop strategies and operate in accordance with impending weather conditions.

The paper is structured as follows: Section 2.1 provides an overview of federated learning and its use cases in weather applications. Section 2.2 covers deep imbalanced learning and implementation details of the classifier. Section 2.3 discusses data augmentation methods and related work, followed by a detailed description of SMOTE and GANs as resampling methods for imbalanced learning. We explore a variety of GANs-based data augmentation methods and provide the outline for implementing centralized and federated learning models. Section 3 presents the experimental results and discussion of five data augmentation methods used for balancing and classifying weather data. In section 4, we provide recommendations for future research problems as a conclusion to our paper.

## 2 Preliminaries

### 2.1 Federated Learning

Federated learning is a collaborative framework where a shared model is trained on a network of decentralized data sources. A copy of the global model is sent to the devices to begin the learning process. The model is trained individually on local datasets, and the model weights are sent back to the server. The weights from multiple centers are aggregated at the global server, and the updated global model is sent back to the centers for further training. This iterative process continues until convergence. The local model is evaluated locally on the validations sets. The global model is evaluated using test data, and the performance metrics at each round are reported.

What differentiates federated learning from other learning algorithms is its ability to maintain privacy. By avoiding the need to store data in a single location, we can harness multiple data sources, safeguard individual privacy, and reduce data storage expenses while achieving high accuracy levels [25, 26]. Access to a large network of data from sources scattered over multiple data centers benefits the deep learning models [27, 28]. Class imbalance appears in many centralized and federated machine learning problems. Xiao et and Wang [29] conducted an experimental study of class imbalance and global performance in federated learning. Due to the variability of local data distribution among all devices and lack of control over client selection, a class imbalance issue arises. This results in a slow convergence rate of the global model. To address this issue, Yang et al. [30] proposed an estimation algorithm that can reveal the class distribution without the need to access the distributed data.

Federated learning has been examined for tackling the issue of imbalanced data in multiple settings. Numerous studies have explored federated learning for problems regarding meteorology and agriculture. Manoj [31] applied federated learning to predict agriculture production using weather data, soil data, and crop management data collected from numerous data silos. Their model improves scalability and ensures privacy, which is essential for users of farming devices. Farooq et al. [32] proposed a federated learning model using Long Short-term Memory (LSTM) neural networks to predict flood, outperforming traditional LSTM models.

In this work, we propose using a deep imbalanced learning model to classify weather data stored in 9 weather stations across Australia. We examine the outcome of training the data centralized and federated. In a centralized learning approach, the stations collect and store their data individually, upon which the model is trained. While this conventional method has its merits, particularly in data consistency and straightforward implementation, it neglects potential issues related to smaller sample sizes and lack of adequate information, especially in imbalanced data. Federated learning is a potent alternative, especially in contexts where data privacy, minimized data transfer, and localized learning are essential. Our experiments compare the two approaches and provide insights into their effectiveness in addressing the challenges posed by data privacy, transfer costs, and geographical variations in weather patterns.

## 2.2 Deep Imbalanced Learning

Deep learning is a network of fully or partially connected neurons organized in layers. The neurons receive the information, and activation functions determine the output of the layers and the network. Deep learning architecture varies according to the specific problem, data structure, the network's depth and size, the layers' functionality, and the optimizer. Schmidhuber [33] conducted a comprehensive historical survey of deep learning and its evolution. Johnson and Khoshgoftaar [34] reviewed the existing methods for the issue of imbalanced data in deep learning. Deep learning is a powerful solution to various real-world problems, and when enhanced by other heuristic feature selection and resampling approaches, it can be very effective for imbalanced learning[35]. Bao et al. [36] introduced a deep learning framework to balance the data in a deeply transformed latent space. In this model, feature learning, balancing, and discriminative learning are conducted simultaneously, which has performed effectively on multi-classification problems. Deep imbalanced learning models are capable of learning imbalanced data in the presence of noise and outliers.

Deep learning has remarkable benefits and has been successful in many classification tasks. The advances in artificial intelligence, particularly deep learning, allow us to create robust models for analyzing diverse data types and provide valuable insights. Therefore, it is selected as the classification model in this study. The architecture of the neural network is problem-specific. We used a sequence of dense layers of various sizes. In federated learning, a layer of Gaussian noise is inserted in between as a hidden layer. The added noise guarantees privacy protection under differential privacy [37] and increases generalization. The standard deviation of the Gaussian noise must be tuned along with other parameters. We removed the noise layer in the architecture to classify the data in a centralized setting. The code snippet for implementing the deep learning model in federated learning is presented in Listing 1.

**Listing 1**  Classifier

```python
# Imports
from keras.layers import GaussianNoise

# Define the model architecture
model = Sequential([
    layers.Dense(256, activation='relu', input_shape=(13,)),
    layers.GaussianNoise(stddev),
    layers.Dense(128, activation='relu'),
    layers.Dense(64, activation='relu'),
    layers.Dense(32, activation='relu'),
    layers.Dense(1, activation='sigmoid')])
```

The batch size is 64, the learning rate is 0.001, and the momentum is 0.9. The standard deviation of the Gaussian noise layer in federated learning architecture is set to 0.01. The model is trained using a Stochastic Gradient Descent optimizer for 30 epochs, and the binary cross-entropy loss function is used. In a federated learning framework, the training is performed in multiple communication rounds, where each round involves the clients sending their model weights to the central server for aggregation. Our experiments suggest that increasing the communication rounds doesn't significantly affect the performance; therefore, the global model is updated for ten communication rounds. In addition to loss and accuracy, the imbalanced learning metrics AUC and G-mean are measured to evaluate the local and global performance. The classification model is implemented using TensorFlow and Keras API in Python. The models are executed in Google Colab Pro with a high-RAM run-time setting.

## 2.3 Data Augmentation

Unlike traditional resampling methods that remove or replicate the existing data points to balance the data, data augmentation is a data processing approach that artificially increases the amount of data by generating synthetic data points from existing data. Resampling methods follow two strategies: removing instances from the majority class (random under-sampling) [38, 39] and adding new instances to the minority class (random over-sampling)[40]. Data augmentation is a set of advanced resampling techniques that focus on generating new instances rather than replicating or removing the original data. Controlling the number of generated samples improves the imbalance ratio and promotes diversity in the data. Since the samples are generated in the feature space, creating a new sample in a nonlinear space improves the results after resampling the minority class. The performance and effectiveness of the data generation techniques are extensively studied by Goel et al. [41], and the limitations of such methods are investigated by Fernandez et al. [42]. Data augmentation methods have been effectively utilized for fraud detection [43], malware and bug report [44], healthcare and medical diagnosis [45, 46], and fault detection in manufacturing and machinery [47]. This paper investigates data augmentation for balancing tabular data using variations of GANs models and the SMOTE, two well-known data augmentation approaches. Figure 3 presents a high-level description of the algorithms.

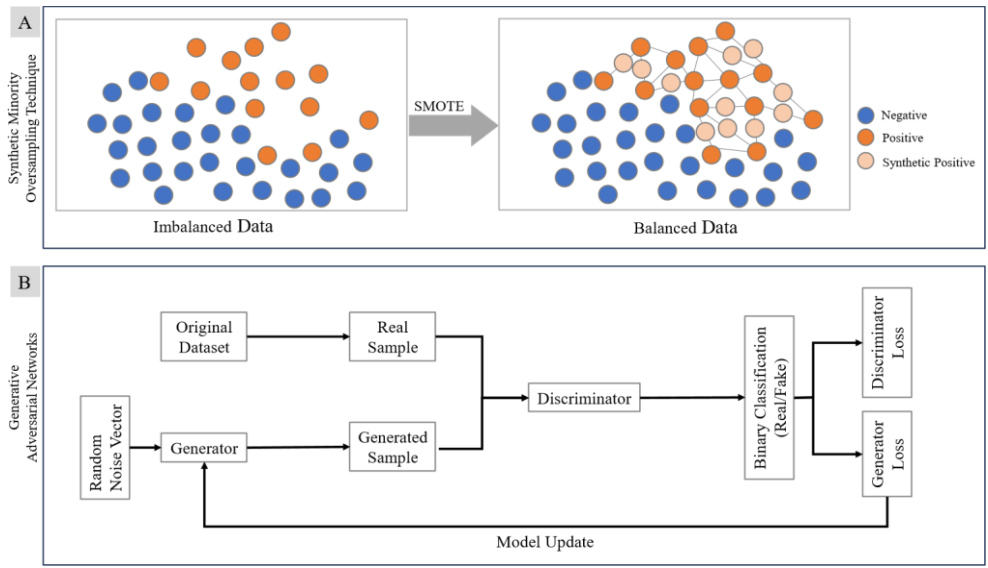

**Fig. 3**: SMOTE and variants of GANs expand the sample size by generating new instances of both classes or balancing the data when only generating instances of the minority class.

Expanding the input data by introducing the samples that represent the original data, combined with the appropriate learning algorithms, strengthens the classification process to attain accurate results for both classes.

### 2.3.1 Synthetic Minority Over-sampling Technique

Over-sampling methods are widely used in imbalanced learning to adjust the data distribution before classification[48, 49]. SMOTE and its variations are among the most popular oversampling methods for tabular data. Chawla et al. [50] combined SMOTE and AdaBoost to enhance training performance by focusing on the most misclassified examples. Borderline-SMOTE starts by identifying the decision boundary between the two classes and then generating samples along the borderline [51]. DeepSMOTE is a novel approach that uses a deep neural network to select the features suitable for generating data with the modified SMOTE algorithm. In SMOTE, the number of instances in the minority class is increased by syntactically creating new instances instead of replicating the existing instances. As shown in Figure 3-A, the new instances are generated based on their nearest neighbors in the feature space. The new examples are added near the line segment that joins the nearest neighbors of the samples in the minority class [52]. Deep imbalanced learning with SMOTE effectively improves G-mean and AUC [53]. Imb-learn Python library [54] is a practical tool for implementing SMOTE.

### 2.3.2 Generative Adversarial Networks

Introduced by [55], GANs is an artificial intelligence scheme designed for learning the underlying patterns in the data in unsupervised learning tasks. Generative models utilize the statistical properties of the data and generate new data points by understanding the data distributions through adversarial learning. Adversarial learning is a machine learning mechanism where two networks with competing objectives are trained simultaneously. Figure 3-B presents the high-level architecture diagram of GANs models. The term adversarial networks refers to the two competing neural network architectures in GANs known as *Generator* and *Discriminator*. The two networks are trained independently. First, the generator takes a vector of random values with Gaussian distribution (random noise vector) and creates a set of new samples. Then, the discriminator takes a sample of original and newly generated data as input and attempts to successfully distinguish between fake and original data in a binary classification problem. The iterative game between the two networks continues until the generator can trick the discriminator into failing to identify the fake data from the original data. The generator and discriminator are trying to optimize their loss function in this setting. Let $G$ be the generator, $D$ be the discriminator, and $z$ be the vector of Gaussian noise fed into the generator. The generator loss function defined in Equation 2.1 measures the binary cross entropy between the output of the discriminator for classifying real and generated data labeled as real.

$$-log(D(G(z))) \quad (2.1)$$

The generator tries to create samples as close as possible to real data. The discriminator loss function defined in Equation 2.2 measures the binary cross entropy for outputs of the discriminator labeled both as real and fake for the generated and real data.

$$-log(D(x)) - log(1 - D(G(z))) \quad (2.2)$$

The discriminator tries to classify real and fake data points correctly. In this two-player game, the two-objective optimization problem is defined as a minimax game with the loss function presented in equation 2.3. The iterative learning process helps us reach the Nash equilibrium between the two networks of $D$ and $G$.

$$min_G max_D V(D,G) = E_x[log(D(x))] + E_z[log(1 - D(G(z)))] \quad (2.3)$$

The competition between the two components guides the generator to create artificial points close to the original data that can not be distinguished from the original dataset. The algorithm follows ten steps:

1. Prepare the dataset, including data cleaning, feature engineering, and normalization of input variables.
2. Define the architecture of the generator and discriminator models.
3. Generate the noise vector and synthesize a sample of fake data.
4. Train the discriminator model on a subset of real and fake samples.
5. Clip the discriminator weights to improve stability
6. Freeze the discriminator weights
7. Train the generator using the output from the discriminator as feedback and generating synthetic data.
8. Iteratively train the GANs model by combining the generator and the discriminator in an adversarial process.
9. Evaluate the GANs model on the validation set.
10. Save the trained generator model and generate synthetic samples to balance the minority class.

Choosing the appropriate architectures and tuning the parameters is one of the main challenges of GANs. The code snippets provided in Listings 2 and 3 are the network architectures used in this study. The batch size is 64, the latent dimension is 13, and the models are trained for 100 epochs. We used Adam optimizer and binary cross-entropy for the loss function.

**Listing 2** Generator

```
# Define the model architecture
model = Sequential([
    layers.Dense(512, activation='relu', input_shape=input),
    layers.Dense(256, activation='relu'),
    layers.Dense(128, activation='relu'),
    layers.Dense(1, activation='sigmoid')])
```

**Listing 3** Discriminator

```
# Define the model architecture
model = Sequential([
    layers.Dense(128, activation='relu', input_shape=input),
    layers.BatchNormalization(),
    layers.Dense(64, activation='relu'),
    layers.Dense(1, activation='sigmoid')])
```

Weight clipping is also used for the discriminator network as a regularization technique. This ensures that the magnitude of the weights is within a predefined range. This technique prevents oscillations and improves the algorithm's stability during training. Numerous studies have explored the use of generative models for handling imbalanced data. GANs have demonstrated outstanding potential in generating data and expanding the sample size with high-quality data close to the original distribution for imbalanced learning problems [56–61]. Divovic et al. [62] improved the quality of generated samples by providing class label context to the network, and Cho and Kim [63] proposed a genetic algorithm approach to find the optimal combination of imbalanced ratios for implementing GANs and SMOTE. Data augmentation using capsule adversarial networks is also a novel approach that constructs a 2-stage model to generate data and then evaluate the balanced dataset by training a classifier. This ensures that the generated data is of good quality [64, 65]. Choi et al. [66] developed a collaborative framework between the generator and classifier to expand the minority sample size and balance the data gradually. GANs data augmentation algorithms have examined a variety of data such as image [67, 68], and tabular datasets for fraud detection, cancer diagnosis, or weather prediction,[69, 70]. In weather applications, GANs generate weather images using a twostep approach where the data is generated and then classified using an ensemble model [71]. Combining GANs with different learning frameworks and preprocessing methods, such as SMOTE, offers promising potential for real-world applications [72].

### Conditional GANs

Conditional GANs (CGANs) are an extension of GANs models that are most effective when the generated data is meant to be tailored to the labels or other class conditioning of the input [73]. In conditional GANs, the conditioning variable (label) is fed as an additional parameter into the generator along with the noise vector. The extra information leads the network to produce data corresponding to the label [74]. Conditional GANs are effective in generating detailed and highly accurate images for supervised and semi-supervised learning problems [75–77].

### Wasserstein GANs

One of the limitations of traditional GANs is their instability during training, which can lead to mode collapse and generate low-quality samples. Getting stuck with a limited range of samples and lacking diversity negatively affects the dataset. Regularization methods and modifications of GANs have been investigated to mitigate this issue [78]. Wasserstein GANs is a variation of GANs trained with a loss function defined based on the Wasserstein distance. Modifying the loss function improves the stability of the model and the quality of the generated samples [79]. Wasserstein distance is measured by the amount of work required to move mass from one distribution to another. Equation 2.4 presents the Wasserstein distance where $P_r$ and $P_g$ are the real and generated distributions, respectively. $Q(P_r, P_g)$ is the set of all couplings of $P_r$ and $P_g$, and $\gamma$ is the joint probability distribution.

$$W(P_r, P_g) = \inf_{\gamma \in Q(P_r, P_g)} E_{(x,y) \sim \gamma}[||x - y||] \quad (2.4)$$

The generator aims to minimize the distance between the distribution of real and generated data, while the discriminator (often referred to as the critic) tries to maximize the distance. Overall, Wasserstein GANs is a promising approach to address the challenges of generating new images, image-to-image translation, and audio synthesis problems. In this paper, we implement the improved Wasserstein GANs with gradient penalty (WGAN-GP) introduced by Gulrajani et al. [80]. The WGAN-GP uses a gradient penalty to ensure the discriminator's gradients are constrained, ensuring Lipschitz continuity and improving algorithm stability during training. This is the main component that differentiates WGAN-GP from the basic WGANs. The architecture of the WGANs-GP discriminator network is presented in Listing 4.

---

**Listing 4** Wasserstein Discriminator (Critic)

```
# Define the model architecture
model = Sequential([
  layers.Dense(128, activation='relu', input_shape=latent_dim),
  layers.BatchNormalization(),
  layers.Dense(64, activation='relu'),
  layers.Dense(1, activation='Linear')])
```
---

In WGANs-GP, the activation function of the final dense layer is linear. The parameter $\lambda$ is the regularization coefficient, which is often 10, and it is a hyperparameter that determines the weight/importance of the gradient penalty in the overall loss (Wasserstein loss). Assuming that $D_{Fake}$ is the average discriminator output of the generated samples and $D_{Real}$ is the average discriminator output of the real samples, the overall loss function is defined as:

$$Loss_{Discriminator} = D_{Fake} - D_{Real} + \lambda * Gradient\ penalty \qquad (2.5)$$

In this model, critic iterations are 5, which is the number of times the discriminator is trained per single generator training. This is a typical WGAN practice to ensure that the critic is well-trained. The clipping value is 0.01, and the learning rate is 0.0001. The models are trained using the RMSprop optimizer implemented with Keras API.

**Minority Generative Adversarial Networks**

The training process of Minority GANs is similar to the traditional GANs, except the generator is only trained on the minority class data to generate 50% of the synthetic data required to balance the data. The generated samples expand the minority class and balance the data, resulting in an unbiased training model with better classification capabilities.

**SMOTE Generative Adversarial Networks**

SMOTE GANs is another extension of GANs with an extra preprocessing step. SMOTE GANs is a hybrid two-phase approach to improve the quality of SMOTE outcomes [81]. First, SMOTE is applied to the minority class data. Then, the generator takes in a sample of original data and samples created using SMOTE. The discriminator is trained to learn the underlying patterns in the feature space to distinguish between real and generated samples. Incorporating GANs and SMOTE to create new samples introduces diversity into the generated model to create accurate, realistic samples.

## 3 Computation Results

Meteorological stations produce vast amounts of data to classify and predict weather patterns, requiring significant computation and storage resources to build machine-learning models and analyze the data. Federated learning is a potential solution for the scalability of weather data analysis and designing a robust model that can generalize well in the presence of noise and synthetic data. The problem being addressed in this experimental study is summarized in Figure 4.

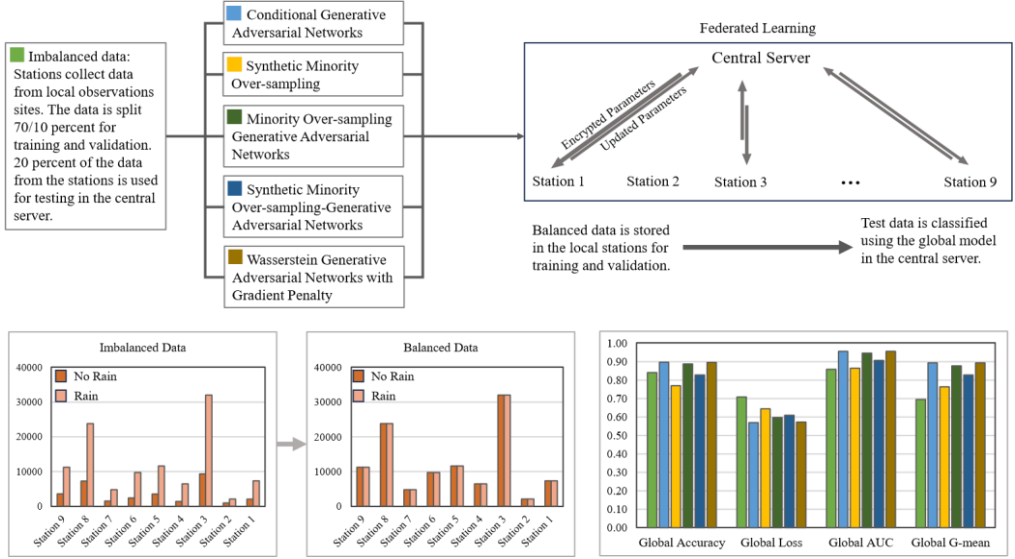

**Fig. 4**: Overview of the federated imbalanced learning problem. We address the issue of imbalanced learning in a federated setting by generating samples of the minority class using 5 data augmentation methods. The local stations train the balanced data, and the encrypted model weights are sent to the global server for aggregation. The results of balanced data training are compared with those of imbalanced data in the federated learning framework.

Initially, each local model is trained on its own imbalanced dataset, recognizing that some weather stations might record rainy days (minority class) less frequently than non-rainy days (majority class). The imbalanced model is used as a baseline to evaluate the data augmentation techniques. We implemented data augmentation strategies, such as SMOTE and various GANs models, to synthesize new instances of the minority class, thereby mitigating the imbalance at each local node. The selected GANs models are CGANs, Minority GANs, SMOTE GANs, and WGANs-GP. The data obtained from the data augmentation methods have equal instances of both classes. This locally balanced data is then utilized to train individual models at each station. The results of training the models in the centralized setting are presented in Figure 5.

To compare the effectiveness of the models, we analyze the performance metrics of each model. Loss is used to measure the error, and lower loss is desired. Accuracy, G-mean, and AUC range from 0 to 1, where higher values are preferred. AUC and G-mean are the most important metrics since they evaluate the models with respect to both classes. Given the results, a summary and analysis based on each augmentation method is provided:

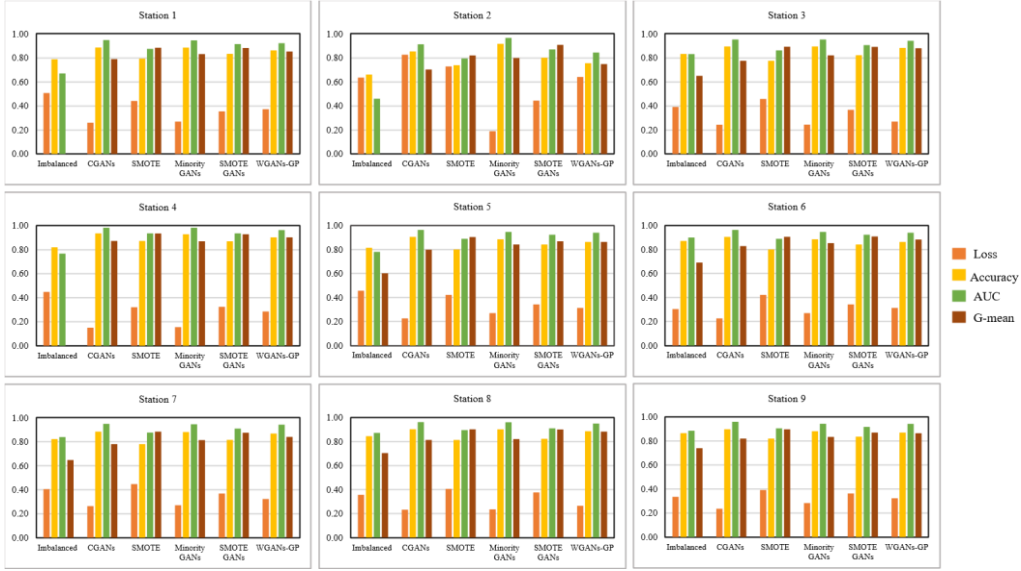

**Fig. 5**: Classification Results of training the models locally on balanced data are compared with the imbalanced data.

1. Imbalanced: This model provides relatively high accuracy and AUC but somewhat lacks in balancing sensitivity and specificity, as evidenced by the lower G-mean, which is expected for imbalanced data.
2. CGANs: CGANs outperform the base model in all stations except 2, where the loss is higher. The overall lower loss in other stations indicates that CGANs have effectively reduced model error. The accuracy and AUC have improved across all stations compared to the imbalanced dataset, highlighting the efficacy of CGANs in distinguishing between classes. The G-mean has also shown significant improvement, showcasing the CGANs' ability to balance sensitivity and specificity.
3. SMOTE: The loss, accuracy, and AUC vary across stations. The accuracy and AUC are similar or slightly higher compared to the imbalanced dataset. However, it has some of the highest G-mean values, suggesting a decent balance between sensitivity and specificity.
4. Minority GANs: Minority GANs have the lowest loss among the models in almost all stations. The loss is lower than the imbalanced dataset across all stations. The accuracy is close to the imbalanced datasets. However, the AUC and G-mean scores are generally very high, which shows a decent balance between sensitivity and specificity. The Minority GANs performed better than CGANs in station 2 while comparable in the remaining stations.
5. SMOTE GANs: Overall, SMOTE GANs perform better than the imbalanced model, with consistently good results across stations. The AUC values suggest good discrimination ability, and the G-mean values indicate a balance between sensitivity and specificity.
6. WGANs-GP: Compared to the imbalanced dataset, WGANs-GP consistently offers competitive or better results across stations. However, the G-mean score is not the highest compared to SMOTE and SMOTE GANs in most stations.

In conclusion, GANs-based augmentation techniques (CGANs, Minority GANs, SMOTE GANs, WGANs-GP) generally outperform the imbalanced datasets regarding all metrics. This suggests that these techniques effectively create synthetic data that aids in better training the models. While a popular method, SMOTE is sometimes surpassed by GANs-based methods, especially in terms of AUC and Accuracy. If one has to rank based on the overall accuracy and AUC across stations, CGANs followed by Minority GANs and WGANs-GP would likely be at the top, and if ranked based on G-mean, SMOTE, and SMOTE GANs would be preferred in most stations.

While the data augmentation techniques prove to be effective in addressing the imbalance ratio between classes, the small sample size impacts the models. Mainly because deep learning models are data-hungry and prefer larger datasets. Federated learning allows us to leverage the power of distributed data while maintaining privacy and security. In the next step, we utilized the balanced data available in the stations to train the federated learning model. The data is horizontally partitioned in the federated learning setup, and the stations collect similar features from the weather observations.

| Model | Accuracy | Loss | AUC | G-mean |
|---|---|---|---|---|
| Imbalanced | 0.841 | 0.71 | 0.859 | 0.695 |
| CGANs | 0.897 | 0.57 | 0.956 | 0.894 |
| SMOTE | 0.770 | 0.64 | 0.865 | 0.764 |
| Minority GANs | 0.888 | 0.60 | 0.947 | 0.877 |
| SMOTE GANs | 0.828 | 0.61 | 0.907 | 0.828 |
| WGANs-GP | 0.896 | 0.57 | 0.956 | 0.844 |

**Table 2**: Classification results of testing the global model trained on the balanced data obtained from the various data augmentation methods, compared with imbalanced data.

The results from evaluating the global model on the test data are presented in Table 2. A brief analysis of the results based on federated learning and data augmentation techniques using the provided metrics is provided.

The imbalanced data is used as the base model for comparison. CGANs and WGANsGP emerge as the superior techniques among the listed, excelling in all metrics. CGANs offer the best G-mean score. This highlights that the conditional generation of synthetic samples can significantly enhance model learning and performance in federated settings. CGANs and WGANs-GP provide reliable and robust synthetic sample generation, thus aiding federated learning models to perform well. Minority GANs also yield very commendable results, with high accuracy, AUC, and G-mean values, though slightly falling short compared to CGANs. Interestingly, SMOTE has reduced accuracy compared to the imbalanced model but shows improvement in loss and a slightly higher AUC and G-mean, indicating an improved balance between sensitivity and specificity but worse overall performance compared to GANs-based models. SMOTE GANs provide improvements in AUC and G-mean compared to the imbalanced model and SMOTE; however, they can not match the performance of other GANs-based models.

In addition to the results presented in Table 2, Figure 6 confirms the stability of the global model over ten communication rounds.

Overall, it appears that the federated learning approach was effective in training a deep-learning model for rainfall prediction. The use of data augmentation methods to balance the dataset improved the accuracy, loss, AUC, and G-mean of the model, highlighting the importance of addressing imbalanced datasets in machine learning. In summary, CGANs and WGANs-GP stand out as particularly effective, achieving the highest performance across all metrics. Minority GANs and SMOTE GANs also enhance performance compared to the baseline (imbalanced) and traditional SMOTE. The purely oversampling-based technique (SMOTE) does not outperform the baseline in terms of accuracy but does improve balance and discrimination between classes (higher G-mean and AUC). Given their very similar performance, the choice between CGANs and WGANs-GP might come down to computational efficiency, storage resources, ease of implementation, and specific use-case requirements. Both methods showcase the potential of GANs-based data augmentation in improving federated learning model outcomes. Analyzing the validation results from local stations allows us to evaluate how effectively the global models generalize and perform on unseen data locally. We'll discuss the results with respect to four metrics: validation loss, validation accuracy, validation AUC, and validation G-mean presented in Figure 7.

In all stations, GANs-based models, particularly CGANs, Minority GANs, and WGANsGP, exhibit lower loss compared to the imbalanced and SMOTE models. Across stations, CGANs and WGANs-GP consistently outperform or match the highest accuracy among the models. They tend to perform very strongly, often achieving AUC above 0.90 and securing top positions in G-mean, suggesting better balance in classifying the majority and the minority classes. Imbalanced models show varied performance across stations and metrics but tend to fall short, especially in AUC and G-mean. Generally, SMOTE does not deliver strong results in loss and accuracy. Its AUC and G-mean are varied, sometimes surpassing imbalanced models but lower than GANs-based models. CGANs, Minority GANs, and WGANs consistently show strong performance in the global metrics and across local stations. They deliver low loss, high accuracy, AUC, and G-mean, which indicates reliable and balanced predictive power for both majority and minority classes. There's visible variability in the performance of SMOTE and imbalanced models globally and locally. They sometimes yield lower accuracy, AUC, and G-mean, indicating difficulty reliably predicting both classes, particularly in imbalanced scenarios. GANs variants indicate strong global robustness, given their top-tier results in global metrics. Their consistent performance across different local stations (despite local data variability) indicates that GANs-based models (especially CGANs, Minority GANs, and WGANs) are not just fitting to the global model but are quite effective locally.

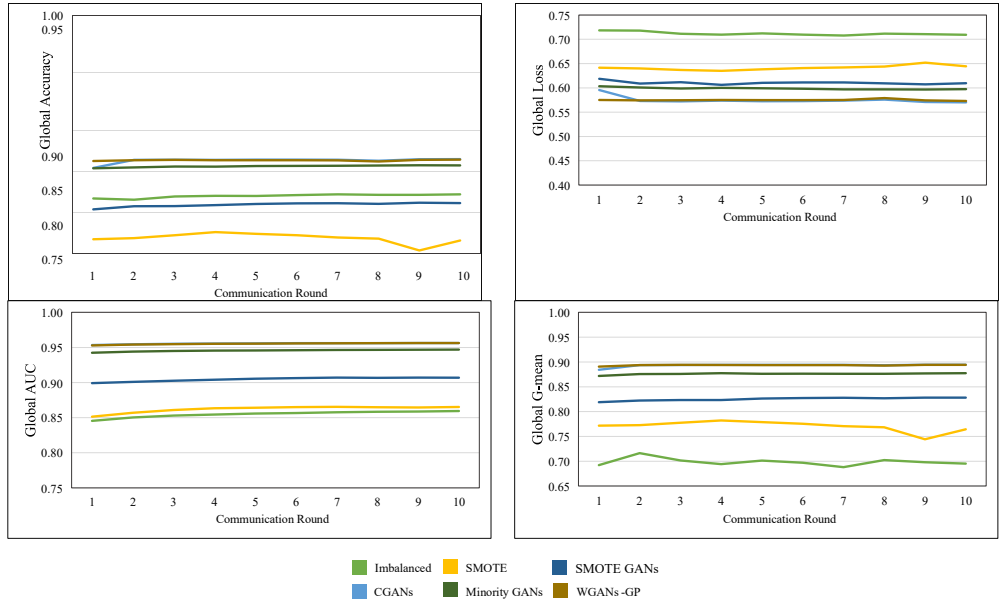

**Fig. 6**: Federated learning process over ten communication rounds, evaluated on global test data.

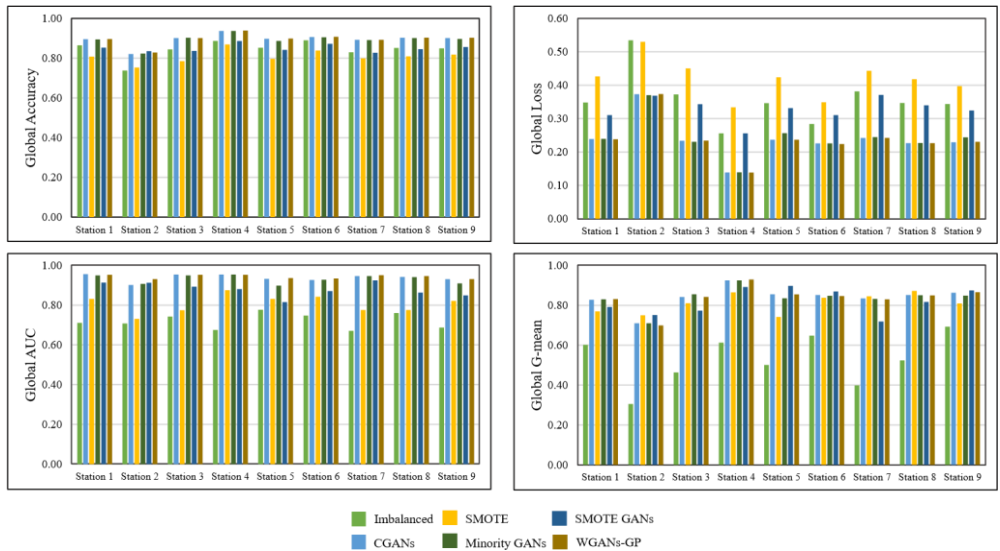

**Fig. 7**: Classification results of federated learning on local validation sets

## 4 Conclusion

This study presents empirical evidence supporting the effectiveness of Generative Adversarial Networks (GANs) models, specifically CGANs, Minority GANs, and WGANs-GP, in navigating the intricacies of federated learning, optimizing the utility of global training and delivering potent performances across local stations.

CGANs, Minority GANs, and WGANs-GP are advanced data augmentation techniques that uniquely contribute to addressing data generation in imbalanced learning scenarios. However, they might potentially degrade model robustness by inadvertently amplifying noise or outliers in the minority class. Implementing GANs variants requires a more intricate design of network architectures in the generator and discriminator and hyperparameter tuning to optimize their capability. Despite offering improved stabilization during training, WGANs-GP can also be computationally demanding due to the implementation of the gradient penalty.

Overall, each of these GANs variants has proven to be theoretically appealing and practically impactful, achieving consistent, robust results in aggregated global metrics and decentralized local validation sets. This experimental study demonstrates the potential of federated learning in meteorology and other climate studies, where data is stored in local stations, and gathering the data in one place is not advised. While imbalanced datasets are a recurring challenge in machine learning, our results suggest combining data augmentation techniques with federated learning can be a viable approach for developing robust and accurate models for predicting weather events.

Our study has several implications for future research. Firstly, the use of federated imbalanced learning can be extended to other applications where data privacy is more constraining, such as individual data available on personal devices used for predicting weather in the user's geographical location. Secondly, further research can investigate different techniques for addressing imbalanced datasets, such as using cost-sensitive learning or class weights in the framework of federated learning. Thirdly, expanding this problem to other weather data formats, such as radar images, is beneficial.

## Declarations

**Conflict of interest** The authors declare no competing interests.
**Consent to participate** Not applicable
**Consent for publication** Not applicable
**Ethics Approval** Not applicable
**Funding** Not applicable
**Data availability** The dataset analyzed during the current study is publicly available in the.
http://www.bom.gov.au/climate/data-services/
**Code availability** The code is available in the author's GitHub repository.
https://github.com/ElahehJafarigol

# References

[1] Haoran Wen, Yang Du, Eng Gee Lim, Huiqing Wen, Ke Yan, Xingshuo Li, and Lin Jiang. A solar forecasting framework based on federated learning and distributed computing. *Building and Environment*, 225:109556, 2022.

[2] Duy-Dong Le, Anh-Khoa Tran, Minh-Son Dao, Kieu-Chinh Nguyen-Ly, Hoang-Son Le, Xuan-Dao Nguyen-Thi, Thanh-Qui Pham, Van-Luong Nguyen, and Bach-Yen Nguyen-Thi. Insights into multi-model federated learning: An advanced approach for air quality index forecasting. *Algorithms*, 15(11):434, 2022.

[3] Caren Marzban and Gregory J Stumpf. A neural network for damaging wind prediction. *Weather and Forecasting*, 13(1):151–163, 1998.

[4] Caren Marzban and Gregory J Stumpf. A neural network for tornado prediction based on doppler radar-derived attributes. *Journal of Applied Meteorology and Climatology*, 35(5):617–626, 1996.

[5] Shaza M Abd Elrahman and Ajith Abraham. A review of class imbalance problem. *Journal of Network and Innovative Computing*, 1(2013):332–340, 2013.

[6] Apurva Sonak and RA Patankar. A survey on methods to handle imbalance dataset. *Int. J. Comput. Sci. Mobile Comput*, 4(11):338–343, 2015.

[7] Theodore B Trafalis, Indra Adrianto, Michael B Richman, and S Lakshmivarahan. Machine-learning classifiers for imbalanced tornado data. *Computational Management Science*, 11(4):403–418, 2014.

[8] Theodore B Trafalis, Huseyin Ince, and Michael B Richman. Tornado detection with support vector machines. In *International Conference on Computational Science*, pages 289–298. Springer, 2003.

[9] Elaheh Jafarigol and Theodore Trafalis. Imbalanced learning with parametric linear programming support vector machine for weather data application. *SN Computer Science*, 1(6):1–11, 2020.

[10] Alberto Fernández, Sara del Río, Nitesh V Chawla, and Francisco Herrera. An insight into imbalanced big data classification: outcomes and challenges. *Complex & Intelligent Systems*, 3(2):105–120, 2017.

[11] Amalia Luque, Alejandro Carrasco, Alejandro Martín, and Ana de las Heras. The impact of class imbalance in classification performance metrics based on the binary confusion matrix. *Pattern Recognition*, 91:216–231, 2019.

[12] David J Hand. Measuring classifier performance: a coherent alternative to the area under the roc curve. *Machine learning*, 77(1):103–123, 2009.

[13] Mohamed Bekkar, Hassiba Kheliouane Djemaa, and Taklit Akrouf Alitouche. Evaluation measures for models assessment over imbalanced data sets. *J Inf Eng Appl*, 3(10), 2013.

[14] César Ferri, José Hernández-Orallo, and R Modroiu. An experimental comparison of performance measures for classification. *Pattern Recognition Letters*, 30(1):27–38, 2009.

[15] Mohammad Hossin and MN Sulaiman. A review on evaluation metrics for data classification evaluations. *International Journal of Data Mining & Knowledge Management Process*, 5(2):1, 2015.

[16] Mehrdad Fatourechi, Rabab K Ward, Steven G Mason, Jane Huggins, Alois Schlögl, and Gary E Birch. Comparison of evaluation metrics in classification applications with imbalanced datasets. In *2008 Seventh International Conference on Machine Learning and Applications*, pages 777–782. IEEE, 2008.

[17] Bartosz Krawczyk. Learning from imbalanced data: open challenges and future directions. *Progress in Artificial Intelligence*, 5(4):221–232, 2016.

[18] T Ryan Hoens and Nitesh V Chawla. Imbalanced datasets: from sampling to classifiers. *Imbalanced learning: Foundations, algorithms, and applications*, pages 43–59, 2013.

[19] Nitesh V Chawla. Data mining for imbalanced datasets: An overview. In *Data mining and knowledge discovery handbook*, pages 875–886. Springer, 2009.

[20] Vaishali Ganganwar. An overview of classification algorithms for imbalanced datasets. *International Journal of Emerging Technology and Advanced Engineering*, 2(4):42–47, 2012.

[21] Guo Haixiang, Li Yijing, Jennifer Shang, Gu Mingyun, Huang Yuanyue, and Gong Bing. Learning from class-imbalanced data: Review of methods and applications. *Expert Systems with Applications*, 73:220–239, 2017.

[22] Alberto Fernández, Salvador García, Mikel Galar, Ronaldo C Prati, Bartosz Krawczyk, and Francisco Herrera. Learning from imbalanced data streams. In *Learning from imbalanced data sets*, pages 279–303. Springer, 2018.

[23] Andrew P Bradley. The use of the area under the roc curve in the evaluation of machine learning algorithms. *Pattern recognition*, 30(7):1145–1159, 1997.

[24] Charles X Ling, Jin Huang, Harry Zhang, et al. Auc: a statistically consistent and more discriminating measure than accuracy. In *Ijcai*, volume 3, pages 519–524, 2003.

[25] Brendan McMahan, Eider Moore, Daniel Ramage, Seth Hampson, and Blaise Aguera y Arcas. Communication-efficient learning of deep networks from decentralized data. In *Artificial intelligence and statistics*, pages 1273–1282. PMLR, 2017.

[26] Tong Li, Zhengan Huang, Ping Li, Zheli Liu, and Chunfu Jia. Outsourced privacypreserving classification service over encrypted data. *Journal of Network and Computer Applications*, 106:100–110, 2018.

[27] Martin Abadi, Andy Chu, Ian Goodfellow, H Brendan McMahan, Ilya Mironov, Kunal Talwar, and Li Zhang. Deep learning with differential privacy. In *Proceedings of the 2016 ACM SIGSAC conference on computer and communications security*, pages 308–318, 2016.

[28] Xiaoguang Niu, Qiongzan Ye, Yihao Zhang, and Dengpan Ye. A privacy-preserving identification mechanism for mobile sensing systems. *IEEE Access*, 6:15457–15467, 2018.

[29] Chenguang Xiao and Shuo Wang. An Experimental Study of Class Imbalance in Federated Learning. In *2021 IEEE Symposium Series on Computational Intelligence (SSCI)*. IEEE, dec 5 2021.

[30] Miao Yang, Ximin Wang, Hongbin Zhu, Haifeng Wang, and Hua Qian. Federated Learning with Class Imbalance Reduction. In *2021 29th European Signal Processing Conference (EUSIPCO)*. IEEE, aug 23 2021.

[31] Inderjeet Mani and I Zhang. knn approach to unbalanced data distributions: a case study involving information extraction. In *Proceedings of workshop on learning from imbalanced datasets*, volume 126, 2003.

[32] Muhammad Shoaib Farooq, Rabia Tehseen, Junaid Nasir Qureshi, Uzma Omer, Rimsha Yaqoob, Hafiz Abdullah Tanweer, and Zabihullah Atal. Ffm: Flood forecasting model using federated learning. *IEEE Access*, 11:24472–24483, 2023.

[33] Jürgen Schmidhuber. Deep learning in neural networks: An overview. *Neural networks*, 61:85–117, 2015.

[34] Justin M Johnson and Taghi M Khoshgoftaar. Survey on deep learning with class imbalance. *Journal of Big Data*, 6(1):27, 2019.

[35] Apurva Sonak, Ruhi Patankar, and Nitin Pise. A new approach for handling imbalanced dataset using ann and genetic algorithm. In *2016 International Conference on Communication and Signal Processing (ICCSP)*, pages 1987–1990. IEEE, 2016.

[36] Feng Bao, Yue Deng, Youyong Kong, Zhiquan Ren, Jinli Suo, and Qionghai Dai. Learning deep landmarks for imbalanced classification. *IEEE Transactions on Neural Networks and Learning Systems*, 2019.

[37] Cynthia Dwork, Aaron Roth, et al. The algorithmic foundations of differential privacy. *Foundations and Trends® in Theoretical Computer Science*, 9(3–4):211–407, 2014.

[38] David A Cieslak, Nitesh V Chawla, and Aaron Striegel. Combating imbalance in network intrusion datasets. In *GrC*, pages 732–737, 2006.

[39] Ch Sarada and M SathyaDevi. Imbalanced big data classification using feature selection under-sampling. *CVR Journal of Science and Technology*, 17(1):78–82, 2019.

[40] Alexander Liu, Joydeep Ghosh, and Cheryl E Martin. Generative oversampling for mining imbalanced datasets. In *DMIN*, pages 66–72, 2007.

[41] Garima Goel, Liam Maguire, Yuhua Li, and Sean McLoone. Evaluation of sampling methods for learning from imbalanced data. In *International Conference on Intelligent Computing*, pages 392–401. Springer, 2013.

[42] Alberto Fernández, Salvador Garcia, Francisco Herrera, and Nitesh V Chawla. Smote for learning from imbalanced data: progress and challenges, marking the 15-year anniversary. *Journal of artificial intelligence research*, 61:863–905, 2018.

[43] Zhenyu Wu, Wenfang Lin, and Yang Ji. An integrated ensemble learning model for imbalanced fault diagnostics and prognostics. *IEEE Access*, 6:8394–8402, 2018.

[44] Xin-Li Yang, David Lo, Xin Xia, Qiao Huang, and Jian-Ling Sun. High-impact bug report identification with imbalanced learning strategies. *Journal of Computer Science and Technology*, 32(1):181–198, 2017.

[45] M Mostafizur Rahman and Darryl N Davis. Addressing the class imbalance problem in medical datasets. *International Journal of Machine Learning and Computing*, 3(2):224, 2013.

[46] Qingyong Wang, Yun Zhou, Weiming Zhang, Zhangui Tang, and Xiaojing Chen. Adaptive sampling using self-paced learning for imbalanced cancer data prediagnosis. *Expert Systems with Applications*, 152:113334, 2020.

[47] Yuan Xie and Tao Zhang. Imbalanced learning for fault diagnosis problem of rotating machinery based on generative adversarial networks. In *2018 37th Chinese Control Conference (CCC)*, pages 6017–6022. IEEE, 2018.

[48] Anjana Gosain and Saanchi Sardana. Handling class imbalance problem using oversampling techniques: A review. In *2017 International Conference on Advances in Computing, Communications and Informatics (ICACCI)*, pages 79–85. IEEE, 2017.

[49] Ruchika Malhotra and Juhi Jain. Handling imbalanced data using ensemble learning in software defect prediction. In *2020 10th International Conference on Cloud Computing, Data Science & Engineering (Confluence)*, pages 300–304. IEEE, 2020.

[50] Nitesh V Chawla, Aleksandar Lazarevic, Lawrence O Hall, and Kevin W Bowyer. Smoteboost: Improving prediction of the minority class in boosting. In *European conference on principles of data mining and knowledge discovery*, pages 107–119. Springer, 2003.

[51] Hui Han, Wen-Yuan Wang, and Bing-Huan Mao. Borderline-smote: a new oversampling method in imbalanced data sets learning. In *International conference on intelligent computing*, pages 878–887. Springer, 2005.

[52] Nitesh V Chawla, Kevin W Bowyer, Lawrence O Hall, and W Philip Kegelmeyer. Smote: synthetic minority over-sampling technique. *Journal of artificial intelligence research*, 16:321–357, 2002.

[53] Damien Dablain, Bartosz Krawczyk, and Nitesh V Chawla. Deepsmote: Fusing deep learning and smote for imbalanced data. *IEEE Transactions on Neural Networks and Learning Systems*, 2022.

[54] Guillaume Lemaîtr̂e, Fernando Nogueira, and Christos K Aridas. Imbalanced-learn: A python toolbox to tackle the curse of imbalanced datasets in machine learning. *The Journal of Machine Learning Research*, 18(1):559–563, 2017.

[55] Ian Goodfellow, Jean Pouget-Abadie, Mehdi Mirza, Bing Xu, David WardeFarley, Sherjil Ozair, Aaron Courville, and Yoshua Bengio. Generative adversarial networks. *Communications of the ACM*, 63(11):139–144, 2020.

[56] Zhuocheng Zhou, Bofeng Zhang, Ying Lv, Tian Shi, and Furong Chang. *Data Augment in Imbalanced Learning Based on Generative Adversarial Networks*, pages 21–30. Springer International Publishing, 2019.

[57] Yangru Huang, Yi Jin, Yidong Li, and Zhiping Lin. Towards imbalanced image classification: a generative adversarial network ensemble learning method. *IEEE Access*, 8:88399–88409, 2020.

[58] Lei Xu, Maria Skoularidou, Alfredo Cuesta-Infante, and Kalyan Veeramachaneni. Modeling tabular data using conditional gan. *Advances in Neural Information Processing Systems*, 32, 2019.

[59] Junhai Zhai, Jiaxing Qi, and Sufang Zhang. Imbalanced data classification based on diverse sample generation and classifier fusion. *International Journal of Machine Learning and Cybernetics*, pages 1–16, 2022.

[60] Tjeng Wawan Cenggoro et al. Deep learning for imbalance data classification using class expert generative adversarial network. *Procedia Computer Science*, 135:60–67, 2018.

[61] Yuxiao Huang, Kara G Fields, and Yan Ma. A tutorial on generative adversarial networks with application to classification of imbalanced data. *Statistical Analysis and Data Mining: The ASA Data Science Journal*, 15(5):543–552, 2022.

[62] Pavle Divovic, Predrag Obradovic, and Marko Misic. Balancing Imbalanced Datasets Using Generative Adversarial Neural Networks. In *2021 29th Telecommunications Forum (TELFOR)*. IEEE, nov 23 2021.

[63] Hwi-Yeon Cho and Yong-Hyuk Kim. A genetic algorithm to optimize smote and gan ratios in class imbalanced datasets. In *Proceedings of the 2020 Genetic and Evolutionary Computation Conference Companion*, pages 33–34, 2020.

[64] Pourya Shamsolmoali, Masoumeh Zareapoor, Linlin Shen, Abdul Hamid Sadka, and Jie Yang. Imbalanced data learning by minority class augmentation using capsule adversarial networks. *Neurocomputing*, 459:481–493, 2021.

[65] M Aslı Aydin. Using generative adversarial networks for handling class imbalance problem. In *2021 29th Signal Processing and Communications Applications Conference (SIU)*, pages 1–4. IEEE, 2021.

[66] Hyun-Soo Choi, Dahuin Jung, Siwon Kim, and Sungroh Yoon. Imbalanced Data Classification via Cooperative Interaction Between Classifier and Generator. *IEEE Transactions on Neural Networks and Learning Systems*, 33(8):3343–3356, 8 2022.

[67] Jia Luo, Jinying Huang, and Hongmei Li. A case study of conditional deep convolutional generative adversarial networks in machine fault diagnosis. *Journal of Intelligent Manufacturing*, 32:407–425, 2021.

[68] Gaofeng Huang and Amir Hossein Jafari. Enhanced balancing gan: Minority-class image generation. *Neural Computing and Applications*, pages 1–10, 2021.

[69] Sankha Subhra Mullick, Shounak Datta, and Swagatam Das. Generative adversarial minority oversampling. In *Proceedings of the IEEE/CVF international conference on computer vision*, pages 1695–1704, 2019.

[70] Justin Engelmann and Stefan Lessmann. Conditional wasserstein gan-based oversampling of tabular data for imbalanced learning. *Expert Systems with Applications*, 174:114582, 2021.

[71] Zhe Li, Yi Jin, Yidong Li, Zhiping Lin, and Shan Wang. Imbalanced adversarial learning for weather image generation and classification. In *2018 14th IEEE International Conference on Signal Processing (ICSP)*, pages 1093–1097. IEEE, 2018.

[72] Shiven Sharma, Colin Bellinger, Bartosz Krawczyk, Osmar Zaiane, and Nathalie Japkowicz. Synthetic oversampling with the majority class: A new perspective on handling extreme imbalance. In *2018 IEEE International Conference on Data Mining (ICDM)*, pages 447–456. IEEE, 2018.

[73] Mehdi Mirza and Simon Osindero. Conditional generative adversarial nets. *arXiv preprint arXiv:1411.1784*, 2014.

[74] Georgios Douzas and Fernando Bacao. Effective data generation for imbalanced learning using conditional generative adversarial networks. *Expert Systems with Applications*, 91:464–471, 1 2018.

[75] Phillip Isola, Jun-Yan Zhu, Tinghui Zhou, and Alexei A Efros. Image-to-image translation with conditional adversarial networks. In *Proceedings of the IEEE conference on computer vision and pattern recognition*, pages 1125–1134, 2017.

[76] Christian Ledig, Lucas Theis, Ferenc Husz´ar, Jose Caballero, Andrew Cunningham, Alejandro Acosta, Andrew Aitken, Alykhan Tejani, Johannes Totz, Zehan Wang, et al. Photo-realistic single image super-resolution using a generative adversarial network. In *Proceedings of the IEEE conference on computer vision and pattern recognition*, pages 4681–4690, 2017.

[77] Han Zhang, Tao Xu, Hongsheng Li, Shaoting Zhang, Xiaogang Wang, Xiaolei Huang, and Dimitris N Metaxas. Stackgan: Text to photo-realistic image synthesis with stacked generative adversarial networks. In *Proceedings of the IEEE international conference on computer vision*, pages 5907–5915, 2017.

[78] Lars Mescheder, Andreas Geiger, and Sebastian Nowozin. Which training methods for gans do actually converge? In *International conference on machine learning*, pages 3481–3490. PMLR, 2018.

[79] Martin Arjovsky, Soumith Chintala, and L´eon Bottou. Wasserstein generative adversarial networks. In *International conference on machine learning*, pages 214–223. PMLR, 2017.

[80] Ishaan Gulrajani, Faruk Ahmed, Martin Arjovsky, Vincent Dumoulin, and Aaron C Courville. Improved training of wasserstein gans. *Advances in neural information processing systems*, 30, 2017.

[81] Anuraganand Sharma, Prabhat Kumar Singh, and Rohitash Chandra. Smotifiedgan for class imbalanced pattern classification problems. *Ieee Access*, 10:30655–30665, 2022.